\documentclass[conference]{IEEEtran}
\usepackage{multirow}
\usepackage{amsmath}
\usepackage{graphicx}
\usepackage{amssymb}    
\usepackage{mathtools}  
\usepackage{bm}         
\usepackage{amsmath}
\usepackage{subfig}    
\usepackage{float}     
\usepackage{epstopdf}
\DeclareMathOperator*{\argmin}{arg\,min}

\begin{document}


\title{Explainable AI for Radar Resource Management: Modified LIME in Deep Reinforcement Learning}

\author{\IEEEauthorblockN{Ziyang Lu, M. Cenk Gursoy,  Chilukuri K. Mohan,  Pramod K. Varshney} 
\IEEEauthorblockA{Department of Electrical Engineering and Computer Science, Syracuse University, Syracuse NY, 13066 
\\
\{zlu112, mcgursoy, ckmohan, varshney\}@syr.edu
}}

\maketitle

\begin{abstract}
Deep reinforcement learning has been extensively studied in decision-making processes and has demonstrated superior performance over conventional approaches in various fields, including radar resource management (RRM). However, a notable limitation of neural networks is their ``black box" nature and recent research work has increasingly focused on explainable AI (XAI) techniques to describe the rationale behind neural network decisions. One promising XAI method is local interpretable model-agnostic explanations (LIME). However, the sampling process in LIME ignores the correlations between features. In this paper, we propose a modified LIME approach that integrates deep learning (DL) into the sampling process, which we refer to as DL-LIME. We employ DL-LIME within deep reinforcement learning for radar resource management. Numerical results show that DL-LIME outperforms conventional LIME in terms of both fidelity and task performance, demonstrating superior performance with both metrics. DL-LIME also provides insights on which factors are more important in decision making for radar resource management.
\end{abstract}


\section{Introduction}

\subsection{Deep reinforcement learning and Cognitive Radar}
Deep reinforcement learning (DRL) is a powerful tool for solving complex decision-making problems in dynamic environments. By integrating neural networks with reinforcement learning, the DRL algorithms can handle complex tasks and yield human-level performance as shown in \cite{lillicrap2015continuous}, \cite{silver2017mastering}, \cite{haarnoja2018soft}, \cite{mnih2015human}.

Cognitive radar is a technology that can enhance radar performance in dynamic and complex environments and recent studies demonstrate the effectiveness of DRL in optimizing the operational parameters of cognitive radars within dynamic environments, such as in radar detection and tracking within congested spectral settings \cite{thornton2020deep}. In \cite{stephan2022scene}, DRL was utilized for scene-adaptive radar tracking. The work in \cite{durst2021quality} introduced a DRL-based approach for quality of service management in radar resource allocation, successfully balancing multiple performance metrics. \cite{9455344} applied reinforcement learning to the revisit interval selection in multifunction radars, where a Q-learning algorithm reduced the tracking load while maintaining a similar track loss probability compared to traditional methods. In our previous studies \cite{10215369} and \cite{ziyang-icc24}, DRL is applied to decide dwell time allocation for multi-target tracking in cognitive radar systems, and  DRL achieves better performance than heuristic approaches under time budget constraints.

\subsection{Local Interpretable Model-agnostic Explanations (LIME)}

As machine learning models become increasingly complex and widely utilized in various fields, an understanding of its decision-making process becomes necessary. Explainable AI (XAI) focuses on interpreting and understanding the model predictions, which is critical for ensuring fairness and enabling model debugging \cite{arrieta2020explainable}. The importance of XAI is particularly evident in high-stakes domains such as healthcare, finance, and autonomous systems, where model decisions can have significant real-world impact \cite{bhatt2020explainable}.

Two approaches have emerged as the leading frameworks for model interpretation: local interpretable model-agnostic explanations (LIME) \cite{ribeiro2016should} and Shapley additive explanations (SHAP) \cite{lundberg2017unified}. While SHAP provides theoretically grounded feature attribution based on cooperative game theory, LIME offers an intuitive and computationally efficient approach by creating local surrogate models to approximate complex model behavior \cite{christoph2020interpretable}, making it more suitable for real-time applications. In this work, we focus on LIME due to its computational efficiency and flexibility, which are crucial for real-time applications in radar systems.

However, LIME's fundamental assumption of feature independence presents a significant limitation in scenarios where features exhibit strong correlations. The original LIME algorithm treats each feature as an independent variable during the perturbation process, potentially generating unrealistic samples that do not align with the underlying data distribution. Recent work has begun that addresses this limitation in various domains. For instance, the work in \cite{shi2020modified} proposes MPS-LIME, a modified perturbed sampling method specifically designed for image classification tasks, which uses graph-based clique construction to maintain spatial correlations between image superpixels. However, this approach is specialized to image inputs and cannot be directly applied to temporal or state-based data like radar systems. To address correlated features, we propose a deep learning assisted LIME (DL-LIME) framework.  

\subsection{Contributions}

In particular, we have the following contributions in this paper. 

\begin{itemize}
    \item In this work, we focus on interpreting the decision-making process of our previously developed DRL approach for radar dwell time allocation.

    \item While LIME offers a promising framework for model interpretation, its fundamental assumption of feature independence becomes limiting in radar systems, where state inputs can exhibit strong correlations due to physical constraints and temporal dynamics. To address this limitation, we propose a modification to LIME that incorporates a deep neural network (DNN) to capture and maintain the correlations between radar state elements, such as target positions and the corresponding tracking costs.

    \item Through comprehensive numerical analysis, we demonstrate that our modified LIME achieves superior performance compared to the conventional LIME approaches in terms of fidelity to the DRL model and task performance. This fidelity measure quantifies how closely the decisions produced by LIME resemble the actual decisions of the DRL model.
    
    \item We validate the effectiveness of our method by providing interpretable explanations of resource allocation strategies of the DRL agent in various radar tracking scenarios.
\end{itemize}

\section{System Model}

We consider a cognitive radar system that dynamically allocates time between scanning for potential targets and tracking detected ones. The system model is based on our previous analysis in \cite{10215369, ziyang-icc24}  with key components summarized below for completeness.

\subsection{Target Motion and Measurement Models}

The target state at time $t$ is defined as $\mathbf{x}_t = [x_t, y_t, \dot{x}_t, \dot{y}_t]^T$, where $(x_t, y_t)$ represents the target's position coordinates and $(\dot{x}_t, \dot{y}_t)$ represents its velocity components. Target motion follows a linear state-space model with Gaussian maneuverability noise:
\begin{equation}
    \mathbf{x}_{t+1} = \mathbf{F}_t \mathbf{x}_t + \mathbf{w}_t
\end{equation}
where $\mathbf{F}_t$ is the state transition matrix and $\mathbf{w}_t$ is the process noise with covariance matrix $\mathbf{Q}_t$, modeling target maneuverability.

The radar obtains range and azimuth measurements $\mathbf{z}_t$ through a nonlinear measurement model with Gaussian noise $\mathbf{v}_t$:
\begin{equation}
    \mathbf{z}_t = h(\mathbf{x}_t) + \mathbf{v}_t = \begin{bmatrix} 
    \sqrt{x_t^2 + y_t^2} \\
    \tan^{-1}(\frac{y_t}{x_t})
    \end{bmatrix} + \mathbf{v}_t
\end{equation}
where the measurement noise $\mathbf{v}_t$ has covariance matrix $\mathbf{R}_t$, and $h(\cdot)$ represents the nonlinear transformation from Cartesian to polar coordinates.

\subsection{Tracking Performance}
Target tracking employs an extended Kalman filter (EKF) to handle the nonlinear measurement model. The tracking performance for each target is evaluated by the following cost function \cite{10215369}:
\begin{equation}
    c_t(\tau_t) = \text{trace}(\mathbf{E}\mathbf{P}_{t|t}\mathbf{E}^T)
\end{equation}
where $\mathbf{P}_{t|t}$ is the posterior state estimate covariance matrix from the EKF, $\mathbf{E}$ is a projection matrix that extracts position components from the state vector, $\tau_t$ is the dwell time allocated to each target within measurement cycle $T_0$. 

\subsection{Scanning and Detection}

The radar's scanning performance is characterized by the signal-to-noise (SNR) ratio:
\begin{equation}
    \text{SNR}_{\text{scan}} = \frac{P_t \tau_{\text{beam}} G_t G_r \lambda_r^2 \sigma}{(4\pi)^3 r^4 L k T_s}
\end{equation}
where
\begin{itemize}
    \item $P_t$ is the transmit power,
    \item $\tau_{\text{beam}}$ is the beam duration,
    \item $G_t$ and $G_r$ are the transmit and receive antenna gains,
    \item $\lambda_r$ is the radar signal wavelength,
    \item $\sigma$ is the radar cross section of the target,
    \item $r$ is the radar-target distance,
    \item $L$ is the system loss factor,
    \item $k$ is Boltzmann's constant,
    \item $T_s$ is the system temperature.
\end{itemize}

The scanning beam duration $\tau_{\text{beam}}$ is related to the total scanning time $\tau_s$ by                                                   
\begin{equation}
    \tau_s = \frac{360^\circ}{\phi}\tau_{\text{beam}}
\end{equation}
where $\phi$ is the phase delay between adjacent radar beams.

The scanning effectiveness is measured by $\Gamma$, defined as the ratio of maximum detectable area to a reference area:
\begin{equation}
    \Gamma = \left(\frac{r_{\text{max}}}{r_0}\right)^2
\end{equation}
where $r_{\text{max}}$ is the maximum detectable range determined by the minimum required SNR for detection, and $r_0$ is a reference range typically set to the default operating range of the radar.

Track initialization follows a 3-of-4 detection model, where three associated measurements within four successive scans are required to establish a track. Measurements are associated with previously received measurements using a global nearest neighbor (GNN) approach with a predefined distance threshold.

\section{Constrained Deep Reinforcement Learning Framework}

Following the approach in \cite{10215369}, we formulate the radar resource management problem for scanning and multi-target tracking as a constrained Markov decision process (CMDP) and solve it using constrained deep reinforcement learning (CDRL). The framework is designed to handle both performance optimization and time budget constraints simultaneously. Instead of the deep Q-network (DQN) based framework in \cite{10215369} and \cite{ziyang-icc24}, we use deep deterministic policy gradient (DDPG) as the algorithm for the DRL framework in this paper.

\subsection{State}
The state $\mathbf{s}_t$ at time $t$ is defined as
\begin{equation}
    \mathbf{s}_t = \left[\left\{\frac{(\hat{x}_{t-1}^n}{\eta}, \frac{\hat{y}_{t-1}^n}{\eta})\right\}_{n=1}^N, \left\{\frac{c_{t-1}^n(\tau_{t-1}^n)}{\eta}\right\}_{n=1}^N, \frac{\lambda_{t-1}}{\eta}\right]
\end{equation}
where $\{(\hat{x}_{t-1}^n, \hat{y}_{t-1}^n)\}_{n=1}^N$ are the estimated locations for all $N$ targets (via EKF) at time $t-1$, $\{c_{t-1}^n(\tau_{t-1}^n)\}_{n=1}^N$ are the tracking costs for all $N$ targets at time $t-1$, $\lambda_{t-1}$ is the dual variable (introduced to manage the time budget constraint as described further below). The size of the state is $3N+1$. $\eta$ is a normalization factor to prevent the inputs to the neural network from being excessively large.  For targets that are not being tracked at time $t-1$, their corresponding position estimates $(\hat{x}_{t-1}^n, \hat{y}_{t-1}^n)$ and tracking costs $c_{t-1}^n(\tau_{t-1}^n)$ are set to zero.

\subsubsection{Action}
The action $\mathbf{a}_t$ is the set of dwell times allocated to the tracking of each target:
\begin{equation}
    \mathbf{a}_t = \{\tau_t^n\}_{n=1}^N, \quad \text{where } \tau_t^n \in [0, T_0]
\end{equation}
The remaining time is allocated to scanning, i.e., $\tau_s = T_0 - \sum_{n-1}^{N} \tau_t^n$.
\subsubsection{Reward}
The reward function $r_t$ is defined as
\begin{equation}
    r_t = U_t(\{\tau_t^n\}_{n=1}^N) - \lambda_t\left(\sum_{n=1}^N \frac{\tau_t^n}{T_0} - \Theta_{max}\right)
\end{equation}
where the first term $U_t(\{\tau_t^n\}_{n=1}^N)$ is the utility function defined as $U_t(\{\tau_t^n\}_{n=1}^N) = -\sum_{n=1}^N c_t^n(\tau_t^n) + \beta \Gamma$, where $\beta$ is the tradeoff coefficient between tracking and scanning performance. The second term is the penalty for violating the time budget constraint, $\Theta_{max}$ denotes the total time budget for tracking, and $\lambda_t$ denotes the dual variable in the Lagrangian relaxation that transforms the constrained optimization problem into an unconstrained one.

In the proposed CDRL framework, the dual variable $\lambda$ is updated simultaneously with the neural network parameters as
\begin{equation}
    \lambda_{t+1} = \max\left(0, \lambda_t + \alpha_\lambda \left(\sum_{n=1}^N \frac{\tau_t^n}{T_0} - \Theta_{max}\right)\right)
\end{equation}
where $\alpha_\lambda$ is the learning rate of the dual variable. The
update of $\lambda_t$ dynamically adjusts the penalty for constraint violations, guiding the learning process toward feasible solutions.

\section{DL-LIME Explanation}
\subsection{LIME}

The framework of local interpretable model-agnostic explanations (LIME) explains the predictions of any classifier or regressor by approximating it locally with an interpretable model. In this paper, we use LIME to explain the decision-making process for the DRL agent discussed in the previous section. The explanation process consists of the following key steps:

\begin{enumerate}
    \item \textbf{Instance Selection:} Given an instance state $\mathbf{s}_t$ to be explained and a trained DRL model $\pi(\cdot)$, LIME first identifies the local region around $\mathbf{s}_t$ where the explanation should be valid.
    
    \item \textbf{Perturbed State Generation:} LIME generates perturbed states around $\mathbf{s}_t$ by sampling from a normal distribution for non-zero continuous features:
    \begin{equation}
        \mathbf{s}_t^{(k)} = \mathbf{s}_t + \mathbf{n}^{(k)}
    \end{equation}
    where $\mathbf{n}^{(k)} = \{n_m^{(k)}\}_{m=1}^{d}$ is the perturbation vector with dimension $d = 3N+1$ (the state space dimension), and $n_m^{(k)} \stackrel{\text{i.i.d.}}{\sim} \mathcal{N}(0, \sigma_m^2)$ for all $d$ components. The parameter $\sigma_m^2$ controls the magnitude of perturbations in the local neighborhood. To better reflect the actual state distribution encountered by the DRL agent, we collect a dataset $\mathcal{D}$ of experienced states by deploying the well-trained DRL agent in diverse radar environments and then compute the empirical mean $\mu_m$ and variance $\sigma_m^2$ for each state component to guide the perturbation sampling process in LIME.
    
    \item \textbf{Similarity Computation:} For each perturbed state $\mathbf{s}_t^{(k)}$, LIME computes a similarity score to the original instance state using an exponential kernel function:
    \begin{equation}
        w(\mathbf{s}_t^{(k)}) = \sqrt{\exp\left(-\frac{D(\mathbf{s}_t, \mathbf{s}_t^{(k)})^2}{a^2}\right)}
    \end{equation}
    where $D(\mathbf{s}_t, \mathbf{s}_t^{(k)}) = \|\mathbf{s}_t - \mathbf{s}_t^{(k)}\|_2$ is the Euclidean distance between the original state $\mathbf{s}_t \in \mathbb{R}^{3N+1}$ and the perturbed state $\mathbf{s}_t^{(k)} \in \mathbb{R}^{3N+1}$, and $a > 0$ is the kernel width parameter controlling the size of the local neighborhood. The similarity score $w(\mathbf{s}_t^{(k)}) \in [0,1]$ decreases exponentially with the distance between states.
    
    \item \textbf{Local Model Training:} LIME fits an interpretable linear model by minimizing the loss function
        \begin{equation}
            \mathcal{L}(\mathbf{w}, b) = \sum_{k=1}^K w(\mathbf{s}_t^{(k)})\|\pi(\mathbf{s}_t^{(k)}) - (\mathbf{w}^T\mathbf{s}_t^{(k)} + b)\|^2 + c\|\mathbf{w}\|^2
        \end{equation}
        with the following terms:
        \begin{itemize}
            \item First term is the weighted mean squared error:
                \begin{itemize}
                    \item $\pi(\mathbf{s}_t^{(k)})$ is the prediction given by DRL agent,
                    \item $\mathbf{w}^T\mathbf{s}_t^{(k)} + b$ is the prediction given by the LIME model,
                    \item $w(\mathbf{s}_t^{(k)})$ weighs errors by similarity to original state.
                \end{itemize}
            \item Second term $c\|\mathbf{w}\|^2$ is a regularization term.
        \end{itemize}
    
    The optimal weights $\mathbf{w}^*$ and bias $b^*$ are found by solving
        \begin{equation}
            (\mathbf{w}^*, b^*) = \argmin_{\mathbf{w}, b} \mathcal{L}(\mathbf{w}, b)
        \end{equation}
    
    \item \textbf{Feature Importance Extraction:} The coefficients $\mathbf{w}^*$ of the fitted linear model provide feature importance scores, indicating how each feature contributes to the model's local prediction.
\end{enumerate}

\subsection{Proposed DL-LIME}
\subsubsection{Motivation}
\begin{figure}
    \centering
    \includegraphics[width=1\linewidth]{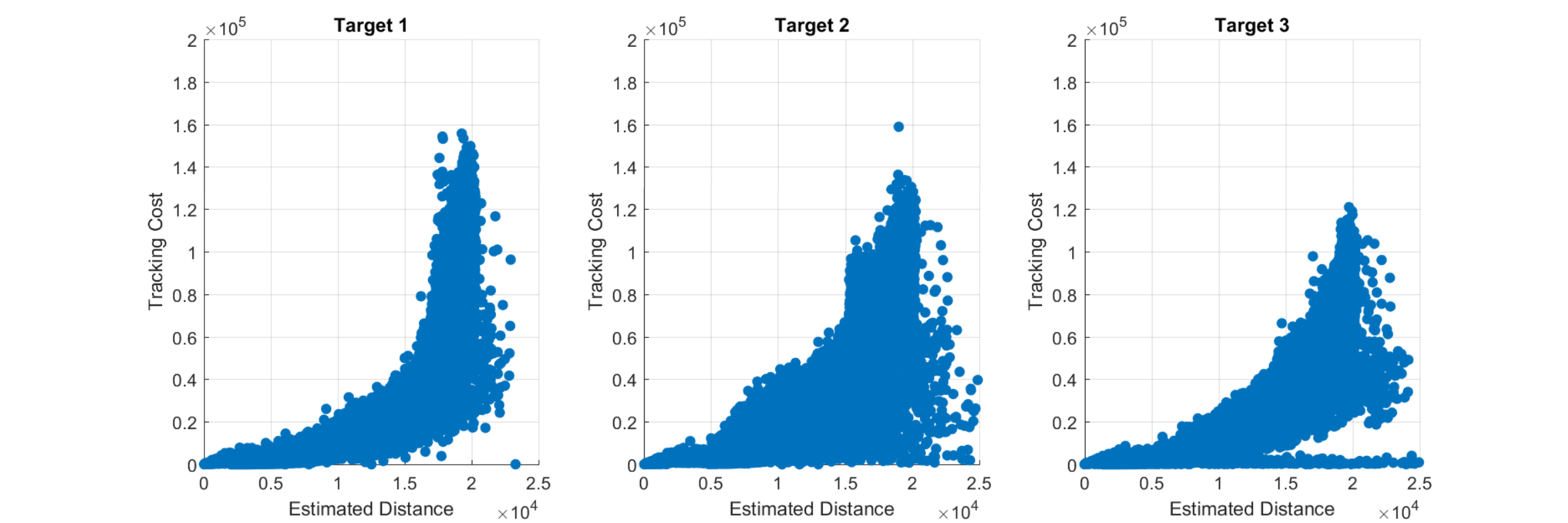}
    \caption{Tracking Costs and Target-to-radar Distances}
    \label{cost_dist}
\end{figure}

The conventional LIME process assumes feature independence during perturbation, which becomes problematic in the considered radar systems where state features exhibit strong correlations. For example, tracking costs exhibit a positive correlation with target distances to the radar. With the same dwell time, tracking targets with larger distances to the radar will be more challenging, resulting in higher costs. To analyze the relationships between state components, we collected trajectory data $\mathcal{D}$ from a well-trained DRL agent operating in the radar environment. For each target $n$, we computed its estimated distance from the radar as $d_t^n = \sqrt{{x_t^n}^2 + {y_t^n}^2}$ and extracted the corresponding tracking costs $c_{t}^n(\tau_{t}^n)$. The scatter plots in Fig. \ref{cost_dist} reveal a strong positive correlation between tracking costs and target-to-radar distances, indicating the inherent relationship between these state components in the considered problem. This correlation suggests that independently perturbing these components during conventional LIME could generate unrealistic states, potentially degrading the reliability of the resulting explanations.

\subsubsection{Implementation}
To address the limitations of the conventional LIME algorithm, we propose an extension that incorporates a deep neural network (DNN) to learn the correlations between state components. The modified LIME algorithm uses this DNN to map selected elements of the state to their correlated counterparts.

As evidenced in Fig. \ref{cost_dist}, tracking costs exhibit strong correlations with other state components. To preserve these inherent relationships during perturbation, we design a DNN that maps the state components to tracking costs. Specifically, the DNN takes $2N+1$ state components as input (excluding tracking costs) and predicts the corresponding tracking costs ${c_{t}^n(\tau_{t}^n)}_{n=1}^N$ for all $N$ targets. This DNN is pre-trained on the collected dataset $\mathcal{D}$ of experienced states to learn the underlying physical relationships.
In our modified LIME approach, rather than independently perturbing all $3N+1$ state components, we only perturb the state components excluding tracking costs, and utilize the trained DNN to generate the corresponding tracking costs. This strategy ensures that the perturbed states maintain physically meaningful relationships between their components, thereby enhancing the fidelity of the local approximation.

\section{Numerical Results}

\subsection{Experimental Setup}

\begin{table}[!t]
\caption{Hyperparameters Used in the Experiment}
\centering
\begin{tabular}{lll}
\hline
Category & Parameter & Value \\
\hline
\multirow{4}{*}{DDPG} & State dimension & 16 \\
& State normalization factor $\eta$ & $10^7$ \\
& Action dimension & 5 \\
& Discount Factor & 0.9 \\
& Action bound (s) & 2.5 \\
& Initial dual variable $\lambda_0$ & 5000 \\
& Step size of dual variable $\alpha_{\lambda}$ & 15000 \\
\hline
\multirow{5}{*}{Environment} & Maximum number of targets $N$ & 5 \\
& Number of Time Slots & 50000 \\
& Required false alarm probability $P_f$ & $10^{-3}$ \\
& Required probability of detection $P_d$& $0.9$\\
& Time budget for tracking $\Theta_{max}$ & 0.9\\
& Target join probability & 0.03 \\
\hline
\multirow{3}{*}{LIME} & Kernel width $a$ & 2.5 \\
& Number of samples & 10000 \\
& Ridge regression regularization parameter $c$ & $10^{-3}$ \\
\hline

\multirow{5}{*}{Radar} 

& Range measurement noise $\sigma_{r,0}^2$ ($m^2$)&16\\
& Azimuth angle measurement noise $\sigma_{\theta,0}^2$ $(\text{rad}^2)$&1e-6\\
& Target maneuverability noise $\sigma_w$ ($(m/s^2)^2$) & 16\\
& Measurement Cycle $T_0$ (s) & 2.5\\
& Tradeoff Coefficient $\beta$ & 100000\\
\hline
\end{tabular}
\label{tab:parameters}
\end{table}

\subsubsection{Hyperparameters}
Hyperparameters are listed in Table \ref{tab:parameters}. In the DDPG algorithm, there are four networks. The actor network and its target network have identical structures, each with two layers containing 256 and 128 neurons, respectively, and ReLU activation functions between the layers.

Similarly, the critic network and its target network share the same structure, consisting of two layers with 100 neurons each, with ReLU as the activation function. Learning rates for both actor and critic networks are set to be 0.0002. 

\subsubsection{Target Spawning Model}
In the experiments, the radar system is placed at the origin and it can track a maximum of $N = 5$ targets. To increase the diversity of the environment and validate the effectiveness of the proposed framework across various conditions, we use the following general target spawning model:

\begin{itemize}
    \item Every 100 iterations, a new target can join the environment with a probability of 0.03. The initial positions and velocities of the targets are generated randomly.
  
    \item To further diversify the operational environment, any target present for more than 3000 time slots will be systematically removed from the environment.
  
    \item It is assumed that the radar is monitoring an area with a radius of 20 km. Targets that move beyond this range are considered to have exited the environment and are no longer tracked by the radar.
\end{itemize}

\subsubsection{DNN in DL-LIME}
A DNN is used for predicting the tracking costs given the inputs as the rest of the state and it is trained on the collected dataset $\mathcal{D}$. The input size is 11 and the output size is 5. There are two feedforward hidden layers, one consists of 128 neurons and the other one consists of 64 neurons. ReLU is used as the activation function and the learning rate is set to be 0.0001.

\begin{figure*}[!ht]
    \centering
    \subfloat[Distances to the Radar of the Targets]{\includegraphics[width=0.32\textwidth]{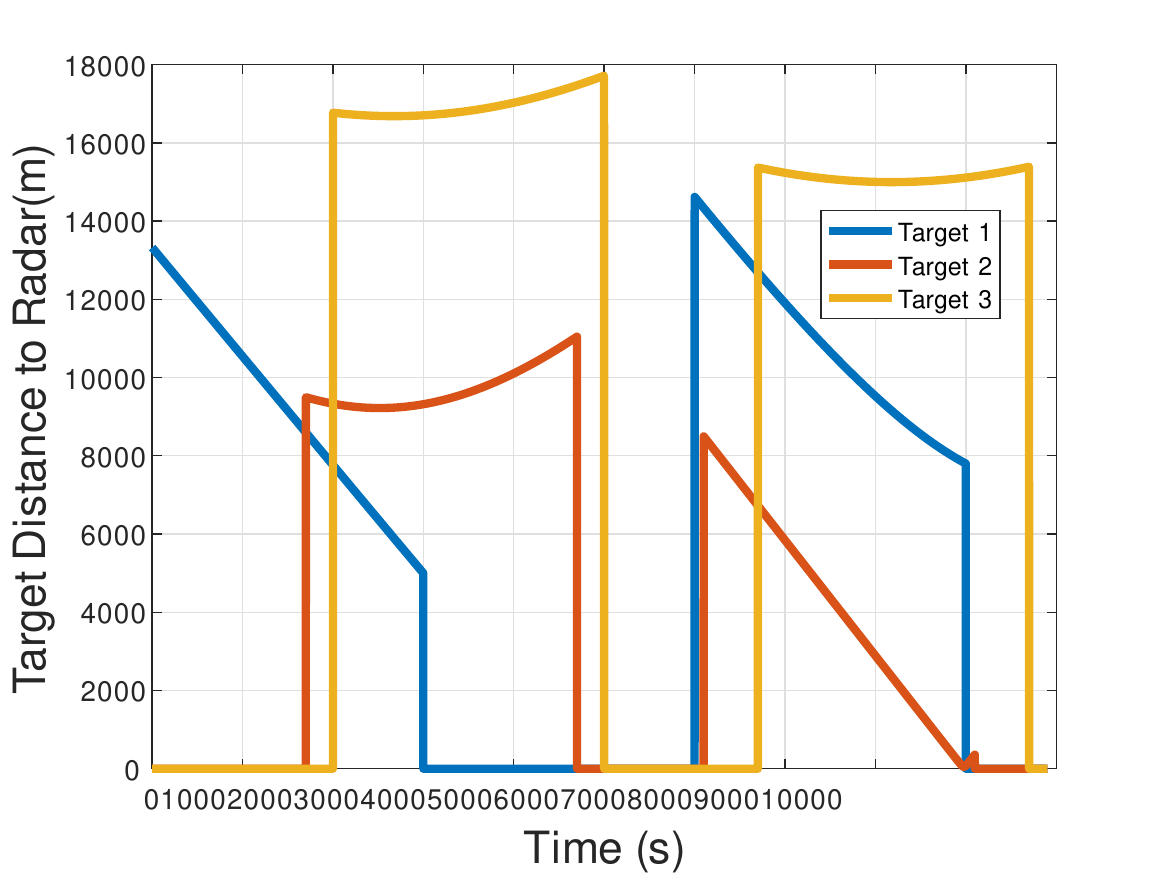}%
    \label{dist}}
    \hfil
    \subfloat[Time Allocation Strategy Predicted by DDPG Agent]{\includegraphics[width=0.32\textwidth]{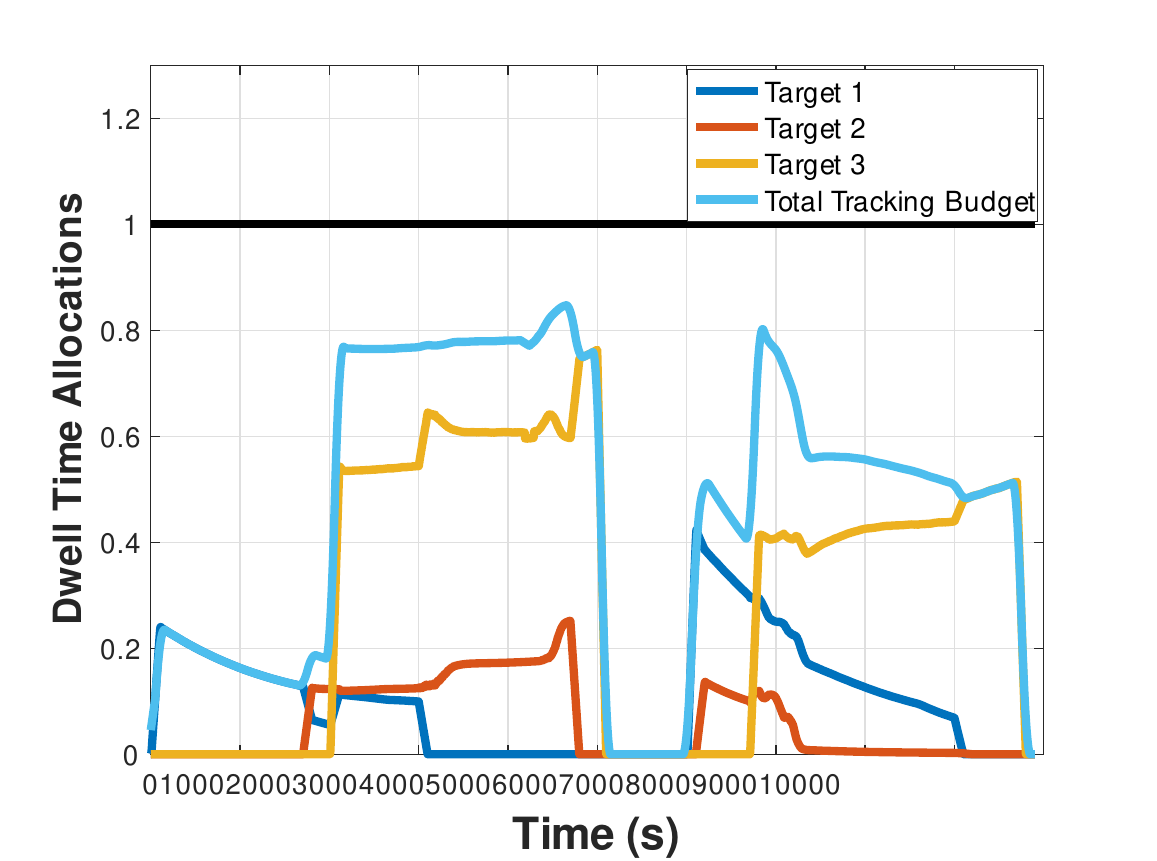}%
    \label{ddpg_act}}
    \hfil
    \subfloat[Time Allocation Strategy Predicted by DL-LIME]{\includegraphics[width=0.32\textwidth]{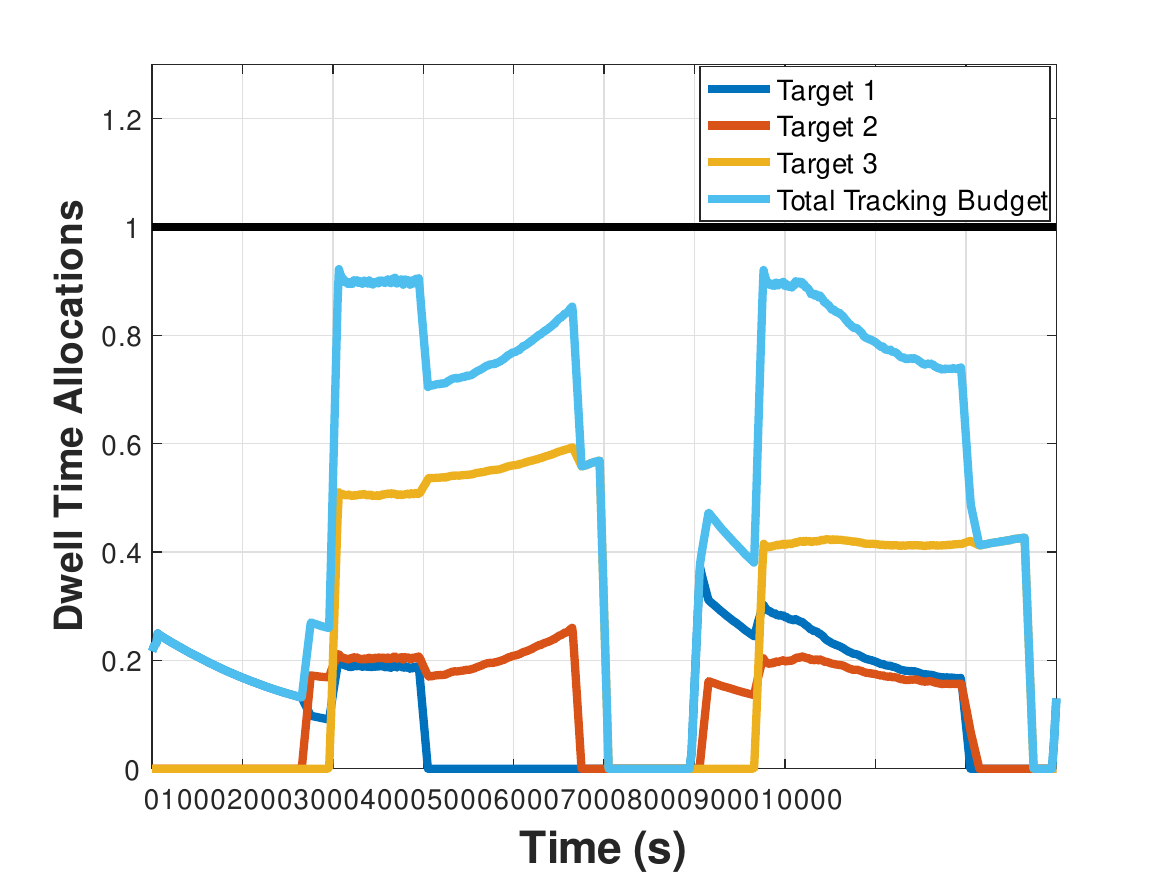}%
    \label{lime_act}}
    \caption{Comparison of DDPG and DL-LIME Time Allocation Strategies}
    \label{action_comp}
\end{figure*}

\begin{figure*}[!ht]
    \centering
   
    \subfloat[Action 2 at $t=4000$]{\includegraphics[width=0.32\textwidth]{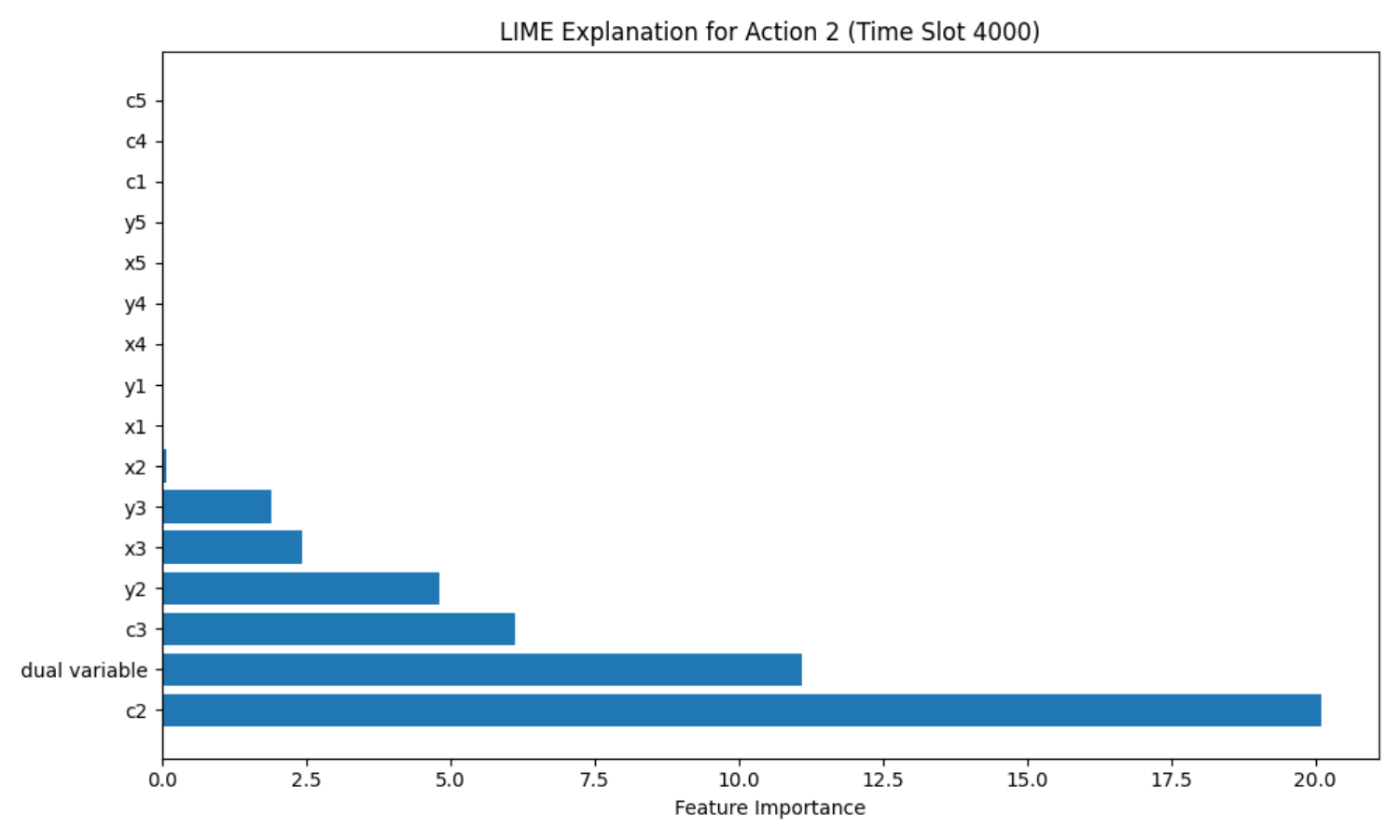}%
    \label{exp_4000_0}}
    \hfil 
    \subfloat[Action 3 at $t=4000$]{\includegraphics[width=0.32\textwidth]{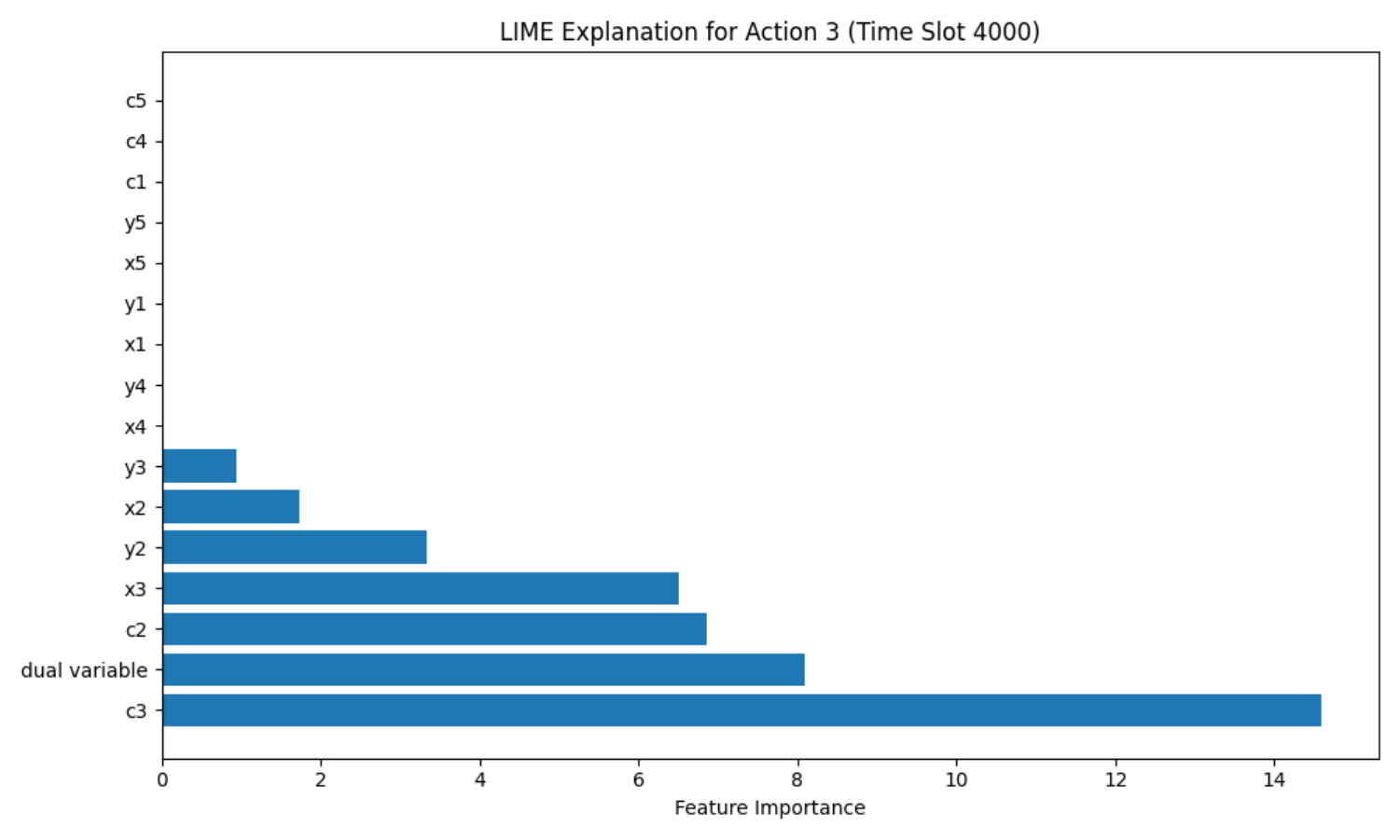}%
    \label{exp_4000_1}}
    \hfil  
    
    \vspace{2mm} 
    
    \subfloat[Action 1 at $t=7000$]{\includegraphics[width=0.32\textwidth]{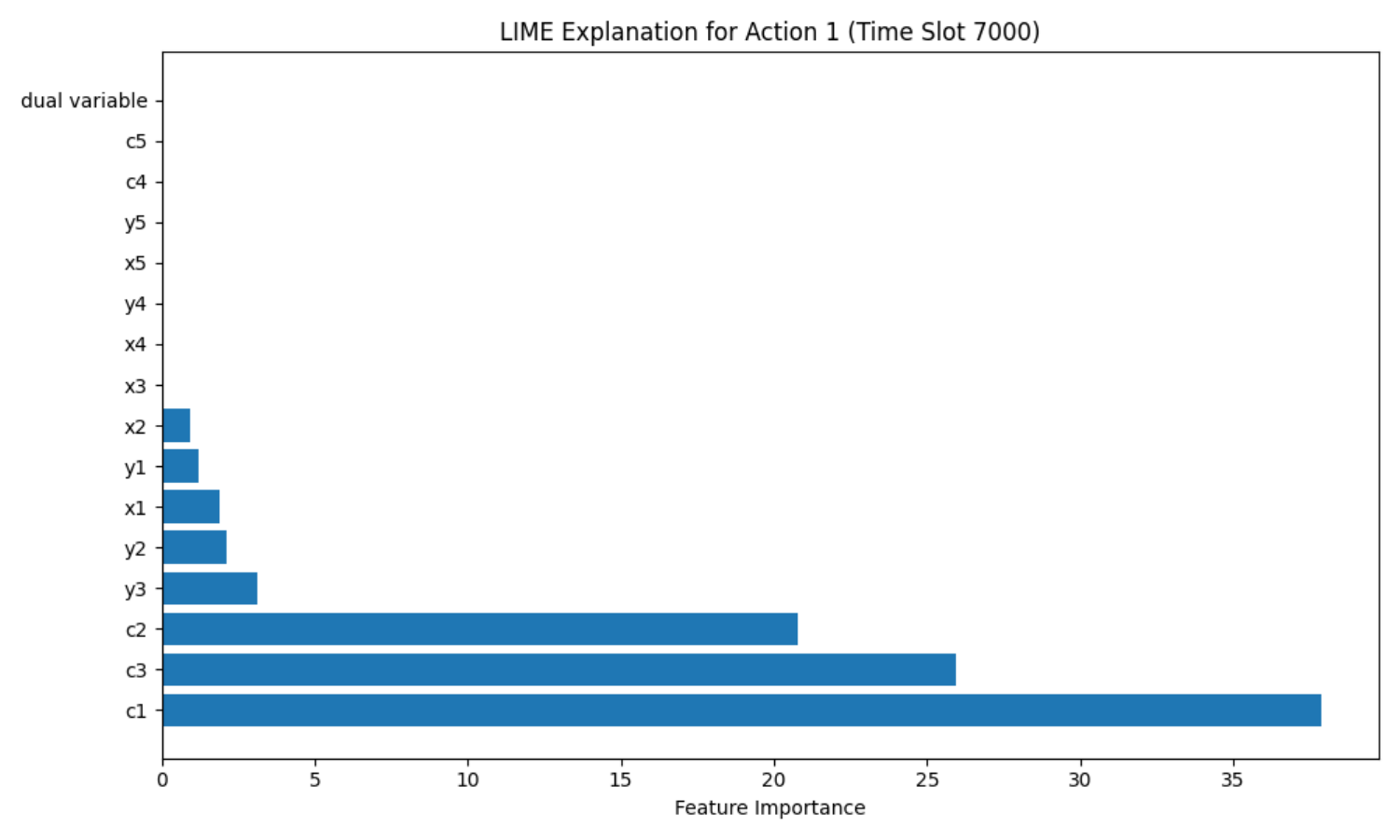}%
    \label{exp_7000_0}}
    \hfil
    \subfloat[Action 2 at $t=7000$]{\includegraphics[width=0.32\textwidth]{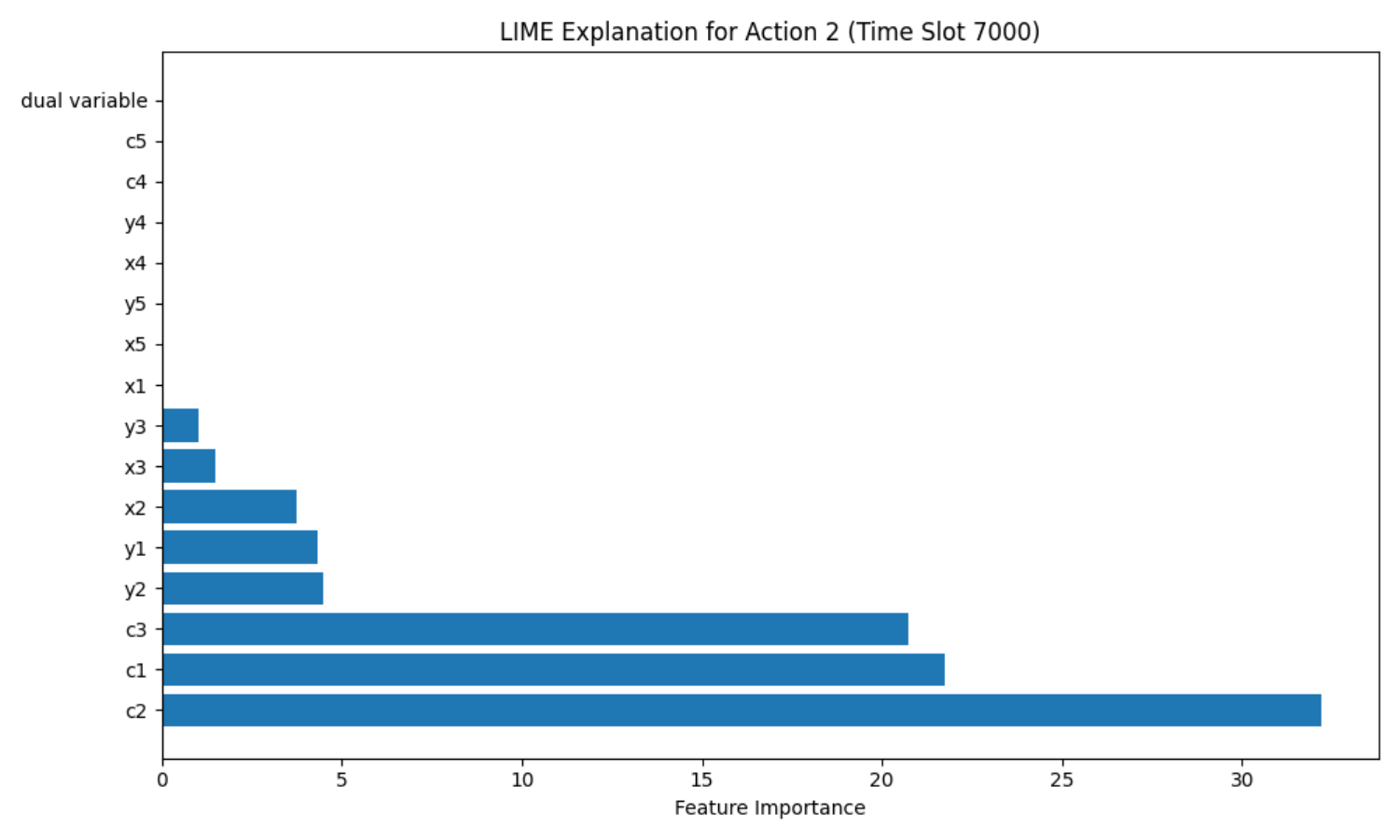}%
    \label{exp_7000_1}}
    \hfil
    \subfloat[Action 3 at $t=7000$]{\includegraphics[width=0.32\textwidth]{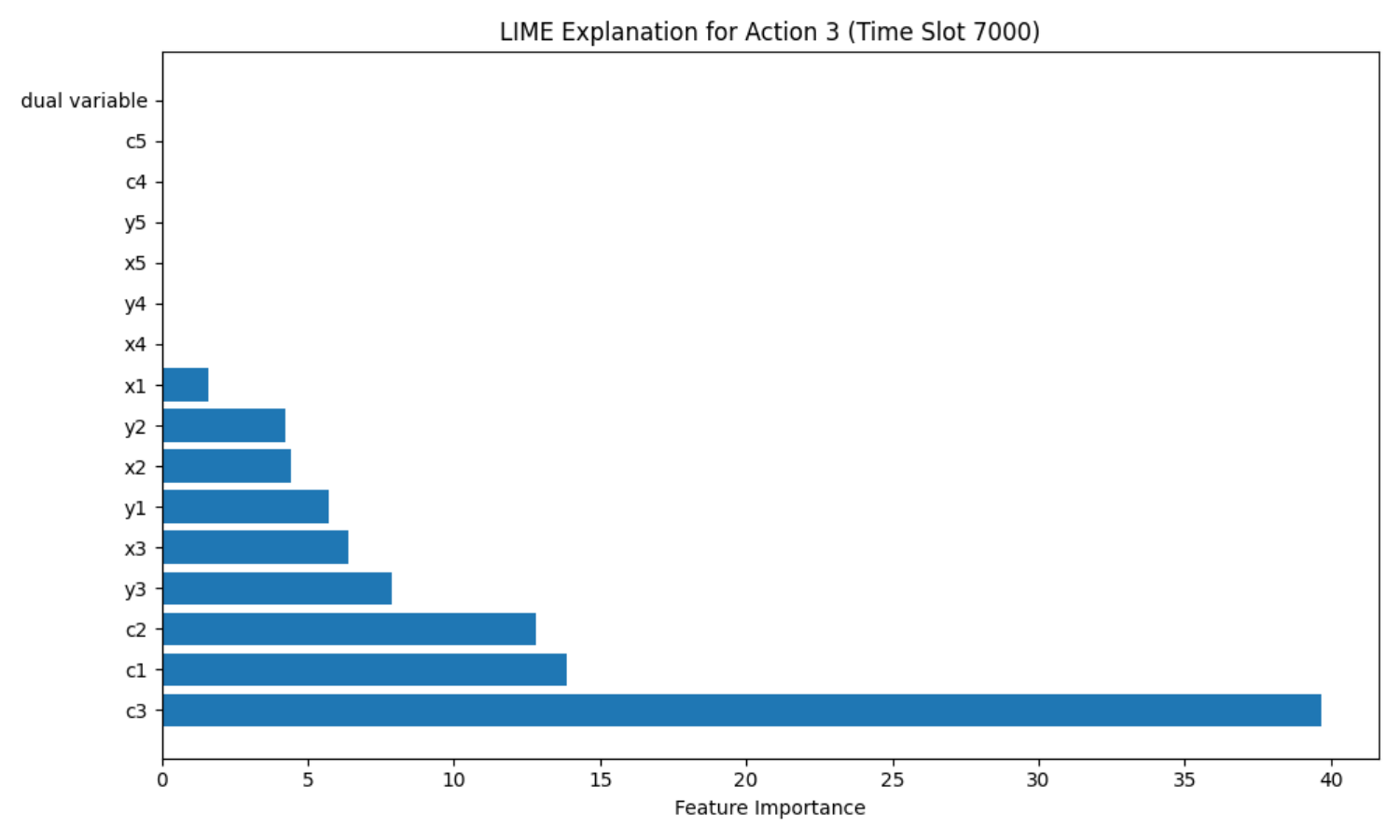}%
    \label{exp_7000_2}}
    \caption{Component importance analysis of DL-LIME explanations at different time points. (a)-(b) show explanations for the actions at $t=4000$, while (c)-(e) demonstrate the explanations for the actions at $t=7000$.}
    \label{lime_explanations}
\end{figure*}

\subsection{Performance Comparison}

To evaluate the proposed framework, we employ three metrics:
\begin{itemize}
    \item \textbf{Average Mean Absolute Error (MAE)}: The MAE quantifies the fidelity of LIME's approximation by measuring the average absolute difference between actions predicted by the LIME model and those chosen by the DDPG agent. We set a checkpoint every 500 time slots. During each checkpoint, a LIME model is found for action explanation. With input as $\mathbf{s}_t$, the actions of the DDPG agent $\mathbf{a}_t$ and the actions provided by the LIME models $\mathbf{a}_t^{LIME}$ can be determined. The MAE is then calculated as
    \begin{equation}
    \text{MAE} = \frac{1}{N} \|\mathbf{a}_{t} - \mathbf{a}_{t}^{\text{LIME}}\|_1.
    \end{equation}
    
    \item \textbf{Average Utility Function}: The utility function evaluates how well LIME models approximate the behavior of the DDPG agent in terms of actual task performance. For each time slot $t$, we find a LIME model to approximate the DDPG policy, then execute the actions predicted by the LIME model in the environment and record the resulting utility function $U_t(\{\tau_t^n\}_{n=1}^N$. This metric provides a practical measure of the ability of LIME to capture the DDPG agent's decision-making effectiveness by directly executing the resulting actions in the environment, rather than just measuring the accuracy of policy approximation as done with the MAE metric.
    
    \item \textbf{Runtime}: The runtime evaluates the computational efficiency by measuring the total runtime required for explaining the decision-making at each time slot.

    \item \textbf{Peak Performance Period}: The peak performance period measures the percentage of time slots in which an algorithm achieves the best performance among all compared methods. Specifically, it represents the fraction of time slots out of 50,000 in which the value of the utility function achieved by the algorithm exceeds the other methods.
\end{itemize}

\begin{table}[htbp]
    \centering
    \caption{Performance comparisons among DL-LIME, LIME, and DDPG}
    \begin{tabular}{|l|c|c|c|c|}
        \hline
        \textbf{Method} & \textbf{MAE} & \textbf{Utility} & \textbf{Runtime (s)} & \textbf{Peak Perf. Period} \\
        \hline
        DDPG & - & $4.39 \times 10^4$ & - & 48.57\%
        \\
        \hline
        LIME & 2.27 & $4.01 \times 10^4$ & 0.42 & 11.20\% \\ 
        \hline
         DL-LIME & 1.95 & $4.49 \times 10^4$ & 1.70 &  40.23\%\\
        \hline
    \end{tabular}
    \label{tab:performance_comparison}
\end{table}

\begin{figure}
    \centering
\vspace{-.4cm}
    \includegraphics[width=0.7\linewidth]{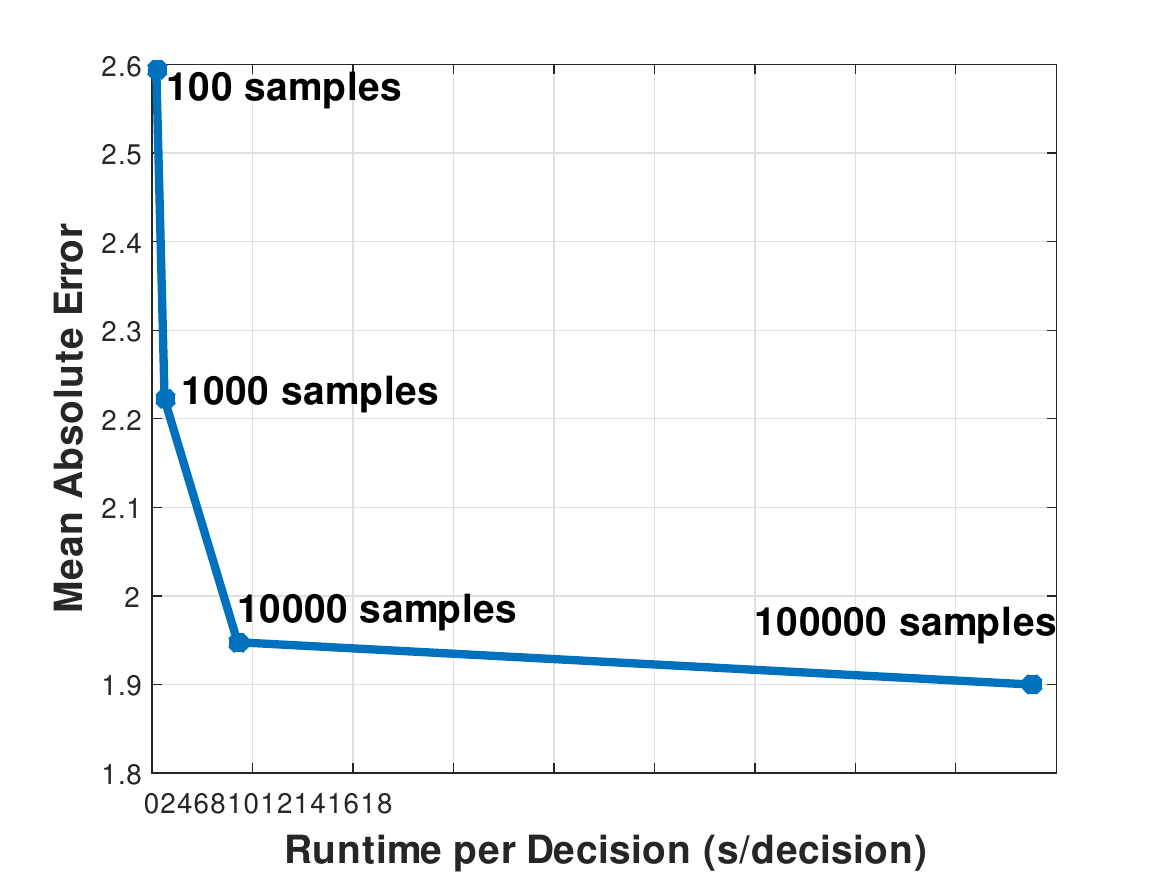}
    \caption{Tradeoff between MAE and Runtime for DL-LIME}
    \vspace{-.3cm}
    \label{tradeoff}
\end{figure}

We have evaluated the average MAE, utility function, runtime (in seconds per decision), and peak performance period of the proposed DL-LIME and conventional LIME approaches, and the performance comparison is provided in Table \ref{tab:performance_comparison}. DL-LIME achieves lower average MAE and higher average utility functions compared to conventional LIME approaches, demonstrating the effectiveness of the proposed DL-LIME approach in approximating the DRL agent's behavior. On the other hand, DL-LIME has larger computational complexity as shown in the runtime comparison due to the incorporation of the deep learning algorithm. DL-LIME results in Table \ref{tab:performance_comparison} are obtained by generating $10,000$ samples for perturbation. By tuning the number of samples in the perturbation process of DL-LIME, one can flexibly strike a balance between computational efficiency and explanation fidelity to meet specific operational requirements. We tested the performance of DL-LIME with different numbers of samples, and the tradeoff between runtime and MAE can be observed in Fig. \ref{tradeoff}.

From the comparison of the peak performance periods, we observe that DDPG outperforms the other methods by maintaining optimal performance for the highest percentage of time slots, followed by DL-LIME which demonstrates superior performance compared to conventional LIME.

\subsection{Actions Predicted by DL-LIME}

With the MAE calculated from the previous section, we can observe that DL-LIME demonstrates higher fidelity to the DDPG policy compared to conventional LIME. In this section, we present the detailed actions predicted by the LIME model. Fig. \ref{dist} plots the distances of the targets to the radar in a testing environment over $10,000$ time slots. Fig. \ref{ddpg_act} and Fig. \ref{lime_act} depict the actions predicted by the DDPG agent and LIME models, respectively. Comparing Fig. \ref{ddpg_act} and Fig. \ref{lime_act}, we observe that the actions predicted by DL-LIME closely resemble those generated by the DDPG agent in the dynamic testing environment, demonstrating the efficiency of the proposed DL-LIME model in approximating the local behavior of the DDPG agent. Both approaches provide reasonable time allocation strategies:

\begin{itemize}
    \item Targets at greater distances from the radar receive higher tracking time allocations, which is evident at $t=4000$. As shown in Fig. \ref{dist}, at this time point, Target 2 and Target 3 are being tracked, with Target 3 at a greater distance from the radar. Both the DDPG agent and the DL-LIME model allocate more tracking time to Target 3. This allocation strategy is intuitive since targets at greater distances experience larger measurement noise, resulting in more challenging tracking conditions and higher tracking costs. Therefore, allocating more of the available time budget to tracking these challenging targets leads to better tracking performance.

    \item When the demand for tracking tasks is low, more time will be allocated to the scanning task, which is evident at $t=1000$ and $t=6500$. At $t=1000$, there is only Target 1 being tracked and hence the tracking demand is low. At $t=6500$, although two targets are tracked, they are relatively close to the radar. In both scenarios, the total tracking time allocation remains low, as shown by the total tracking budget curve. Radar tends to allocate more time for scanning the environment (in order to detect newly emerging targets), which optimizes the overall utility function.
\end{itemize}

\subsection{Explanations}

In this section, we examine the explanation given by the DL-LIME approach in specific time slots. Fig. \ref{lime_explanations} presents the importance of the state components in the explanations of individual action decisions. Note that the action $\mathbf{a}_t = \{\tau_t^n\}_{n=1}^N$ consists of the dwell time allocated for tracking the present targets. Here, we denote the decision $\tau_t^n$ as Action $n$. According to the results in Fig. \ref{lime_explanations}, the LIME explanations reveal a consistent pattern in which the cost  $c_{t-1}^n(\tau_{t-1}^n)$ associated with tracking Target $n$ in the previous time slot holds dominant importance in the decision-making for Action $n$ (i.e., dwell time decision for tracking Target $n$ in the current time slot). For instance, it can be seen that the tracking cost of Target 2 is the factor with the highest importance in the explanation of Action 2 as shown in Fig. \ref{exp_4000_0} and the tracking cost of Target 3 is the dominant factor in the explanation of Action 3 as shown in Fig. \ref{exp_4000_1}. We only show the explanations for Action 2 and Action 3 since only Target 2 and Target 3 are tracked at $t=4000$. 

While $c_{t-1}^n(\tau_{t-1}^n)$ dominates in the explanation of its corresponding Action $n$, the tracking cost components of the other targets also show substantial importance as evidenced in Fig. \ref{lime_explanations}. This suggests that the model considers interactions between targets when making allocation decisions. This explanation also shows that the DDPG agent does not make decisions for all the targets in isolation. Instead, it weighs the tracking costs of all the targets to make a balanced allocation.

\section{Conclusions}

In this paper, we proposed a DL-LIME framework that enhances the conventional LIME approach by incorporating deep learning algorithms in the sampling process in order to take feature correlation into consideration. The performance of the proposed DL-LIME was evaluated against conventional LIME using multiple metrics: fidelity to the neural network, task performance, and computational runtime. Numerical results demonstrate that DL-LIME achieves superior performance in both mean absolute error (MAE) and task-specific metrics in radar resource management. Furthermore, the framework shows strong capability in approximating the local behavior of the DRL agent, as evidenced by the detailed analysis of explanations at specific time instances. These results demonstrate that DL-LIME not only improves the performance of traditional LIME but also provides interpretable insights into the decision-making process of complex DRL algorithms for radar resource allocation.

\bibliographystyle{IEEEtran}
\bibliography{ref}
\end{document}